\documentclass{article}

\usepackage{microtype}
\usepackage{graphicx}
\usepackage{subfigure}
\usepackage{booktabs} 

\usepackage{hyperref}

\usepackage[accepted]{icml2023}

\usepackage{amsmath}
\usepackage{amssymb}
\usepackage{mathtools}
\usepackage{amsthm}
\usepackage{microtype}
\usepackage{listings}
\usepackage{inconsolata}
\usepackage{enumitem}

\lstdefinelanguage{Julia}%
  {morekeywords={abstract,break,case,catch,const,continue,do,else,elseif,%
      end,export,false,for,function,immutable,import,importall,if,in,%
      macro,module,otherwise,quote,return,switch,true,try,type,typealias,%
      using,while,@node,@init,@prev,@observe,@nodecall,rand},%
   sensitive=true,%
   alsoother={\$},%
   morecomment=[l]\#,%
   morecomment=[n]{\#=}{=\#},%
   morestring=[s]{"}{"},%
   morestring=[m]{'}{'},%
}[keywords,comments,strings]%

\def\jl{\lstinline[basicstyle=\normalsize\ttfamily]}

\usepackage[capitalize,noabbrev]{cleveref}

\theoremstyle{plain}
\newtheorem{theorem}{Theorem}[section]
\newtheorem{proposition}[theorem]{Proposition}

\theoremstyle{definition}

\theoremstyle{remark}
\newtheorem{remark}[theorem]{Remark}

\usepackage[textsize=tiny]{todonotes}

\icmltitlerunning{Automatic Rao-Blackwellization for SMC}

\begin{document}

\twocolumn[
\icmltitle{Automatic Rao-Blackwellization for Sequential Monte Carlo\\ with Belief Propagation}



\begin{icmlauthorlist}
\icmlauthor{Waïss Azizian}{grenoble}
\icmlauthor{Guillaume Baudart}{psl}
\icmlauthor{Marc Lelarge}{inria}
\end{icmlauthorlist}

\icmlaffiliation{grenoble}{LJK, Univ.~Grenoble Alpes, France}
\icmlaffiliation{inria}{Inria -- ENS -- PSL University, France}
\icmlaffiliation{psl}{ENS -- PSL University -- CNRS -- Inria, France}

\icmlcorrespondingauthor{Marc Lelarge}{marc.lelarge@ens.fr}

\icmlkeywords{Machine Learning, ICML}

\vskip 0.3in
]



\printAffiliationsAndNotice{}  

\begin{abstract}

Exact Bayesian inference on state-space models~(SSM) is in general untractable, and unfortunately, basic Sequential Monte Carlo~(SMC) methods do not yield correct approximations for complex models.
In this paper, we propose a mixed inference algorithm that computes closed-form solutions using belief propagation as much as possible, and falls back to sampling-based SMC methods when exact computations fail.
This algorithm thus implements automatic Rao-Blackwellization and is even exact for Gaussian tree models.
\end{abstract}

\section{Introduction}

In this paper, we focus on online Bayesian inference for state-space models~(SSM).
A characteristic example is an agent which relies on a tracker model to continuously estimate its position from noisy observations, and use the current estimation to decide its next action.

This work focuses on Sequential Monte Carlo (SMC) inference algorithms \cite{smc} introducing a small probabilistic programming language (PPL) to validate new algorithms. We implemented the simplest SMC method: the bootstrap particle filter \cite{gordon1993novel} requiring only simulation of the prior distribution. While widely applicable, it is suboptimal with respect to Monte Carlo variance in situations where analytical relationships between random variables (such as conjugate priors or affine transformations) can be exploited. Within SMC, this translates into improvements such as Rao-Blackwellization \cite{doucet2009tutorial}, and this paper seeks to automate it for the user of our PPL.

This paper presents the following contributions:
1) A Julia domain specific language \textit{OnlineSampling.jl} to describe SSM focusing on reactive models, i.e., streaming probabilistic models based on the synchronous model of execution. 2) A new algorithm mixing approximate SMC methods with exact belief propagation for online Bayesian inference.

In Section \ref{sec:reac}, we briefly discuss the specificities of our PPL (for general purpose PPL in Julia see \textit{Turing.jl} \cite{ge2018t} or \textit{Gen.jl} \cite{Cusumano-Towner:2019:GGP:3314221.3314642}). In Section \ref{sec:rao}, we present our new algorithm based on belief propagation for the Rao-Blackwellization. Some experiments are presented in Section \ref{sec:expe} and related work in Section \ref{sec:related}. 

The code is available at:\\\url{https://github.com/wazizian/OnlineSampling.jl}.

\section{Reactive probabilistic programming}\label{sec:reac}
To program reactive probabilistic models, we designed a Julia embedded domain specific language inspired by ProbZelus~\cite{probzelus}. 
Following the dataflow synchronous approach~\cite{synchronous_languages}, programs execute in lockstep on a global discrete logical clock. 
Inputs and outputs are data streams, and programs are stream processors.

A stream function is introduced by the macro \jl{@node}. 
Inside a node, the macro \jl{@init} declares a variable as a memory. 
Another macro \jl{@prev} accesses the value of a memory variable at the previous time step (\jl{@prev} macros can be nested to access values arbitrarily back in time).

In the line of recent probabilistic programming languages~\cite{anglican, webppl, birch, pyro}, our language is extended with two probabilistic constructs: 1) \jl{x = rand(D)} introduces a random variable with prior distribution \jl{D}, 2) \jl{@observe(x, v)} conditions the models assuming the random variable \jl{x} takes the value \jl{v}.

\subsection{Running example}\label{sec:ex_code}
As a running example, consider a simple tracker model which continuously estimates the position of a runner on a trail from noisy observations of its speed and altitude and a map of the trail. 
This model can be implemented as follows (a mathematical description is provided in Equation \eqref{eq:model}):

\begin{lstlisting}
@node function model()
  @init s = rand(Normal(0, $\sigma^s_0$))
  @init x = rand(Normal(0, $\sigma^x_0$))
  s = rand(Normal(@prev(s), $\sigma^s$))
  x = rand(Normal(@prev(x) + @prev(s), $\sigma^x$)
  a = alt(x)
  return x, s, a
end

@node function tracker(s_obs, a_obs)
  x, s, a = @nodecall model()
  t = rand(Normal(s, $\sigma^t$))
  b = rand(Normal(a, $\sigma^b$))
  @observe(t, s_obs)
  @observe(b, a_obs)
  return x
end
\end{lstlisting}

The stream function \jl{model} describes the generative model of the runner and returns at each time step $t \in \mathbb{N}$ the current state, i.e., position $x_t$, speed $s_t$, and altitude $a_t$.
The speed follows a Gaussian random walk (Line~4).
The position is Gaussian distributed around the (discrete) integral of the speed (Line~5).
The altitude can then be deduced from the map using the function \jl{alt}, which maps a position to an altitude (Line~6).
All parameters $\sigma$ are constant.

The stream function \jl{tracker} then conditions this model on noisy observations from a speedometer (\jl{s_obs}) and an altimeter (\jl{a_obs}). Line 11 specifies that the position, speed, and altitude are r.v. generated according to the \jl{model}.
We assume that both observations are Gaussian distributed around the estimations computed by the model (Lines~12 to~15).

\begin{remark}\label{re:obs}
    For ease of presentation, we separate the \jl{model} for the generative model and the \jl{tracker} for the inference. As a result, here, observations are made ``after" the model, but we will see later that making observations inside the model can be helpful, and this is doable in our framework.
\end{remark}

\subsection{Mixed inference with SMC}

For such models, inference is a discrete process that returns the posterior distribution of state at the current time step given the observations so far.
Unfortunately, for complex models, there is no closed-form solution for the posterior distribution.
For instance, in our example, the call to the \jl{alt} function makes the problem intractable.
Basic Sequential Monte Carlo~(SMC) methods on the other hand do not yield correct approximations for complex models.

To mitigate these issues, mixed inference algorithms~\cite{delayed_sampling, probzelus, semi-symbolic} extend an SMC sampler with the ability to perform exact computations on subsets of random variables.
The SMC sampler launches $N$ independent simulations of the model, or \emph{particles}.
Each particle performs exact computations as much as possible and, when it fails, samples concrete values for a few random variables before resuming the computations.
These methods thus implement automatic Rao-Blackwellization.

Following these ideas, we propose a new mixed inference algorithm that uses Gaussian belief propagation~\cite{weiss1999correctness} for exact computations.
Inference is thus exact for all models (or parts of a model) that can be expressed as Gaussian Trees.
For instance, in the runner example, it is possible to compute exact distributions for~$s$ and~$x$ and sample $a$ to perform the last observation.

\section{Rao-Blackwellization with belief propagation}\label{sec:rao}
We describe in the next section the subroutine used to do exact marginalization for Gaussian trees thanks to belief propagation and show in the following section how it applies to our dynamic online setting.
\subsection{Belief propagation for Gaussian trees}
Static probabilistic model: we consider a rooted tree $T =(V,E,r)$ where $(V,E)$ is a tree, i.e. an acyclic graph, and $r\in V$ is a particular node of the tree called the root. In a rooted tree, there is a natural notion of parent and children: for each node $v\in V\backslash\{r\}$, there is a unique node in the tree closest to the root in the neighbors of $v$. This node is called the parent of $v$ and denoted by $\pi(v;r)$.
The other neighbors of $v$ in the tree are called the children and are denoted by $c(v;r)\subset V$. We also define the children of the root $c(r;r)$ as the set of neighbors of the root $r$. For $v\in c(r;r)$, we denote by $T_v$ the tree rooted at $v$ obtained when the root $r$ is removed from the original tree $T$.

For a rooted tree $T=(V,E,r)$, we associate with each $v\in V$, a random variable $x_v$ such that the distribution satisfies:
\begin{eqnarray}
\label{eq:def_gt}p((x_v)_{v\in V}) = p(x_r) \prod_{v\in V\backslash \{r\}}p(x_v|x_{\pi(v;r)}).
\end{eqnarray}
A Gaussian tree corresponds to the particular case where $p(x_r)=\mathcal{N}(\mu_r,\Sigma_r)$ is the density of a Gaussian random variable and all conditional probabilities $p(x_v|x_u)$ with $u={\pi(v;r)}$ correspond to linear Gaussian models:
\begin{eqnarray*}
    p(x_v|x_u) = \mathcal{N}(x_v| A_{(v|u)}x_u+b_{(v|u)},\Sigma_{(v|u)}),
\end{eqnarray*}
with fixed matrices $A_{(v|u)}, \Sigma_{(v|u)}$ and vector $b_{(v|u)}$.

For a Gaussian tree, marginalization of the root is straightforward. From \eqref{eq:def_gt}, we see that
\begin{eqnarray*}
p((x_v)_{v\in V}) &=& p(x_r) \prod_{v\in c(r;r)}p((x_u)_{u\in T_v}|x_r)\\
&=& p(x_r)\prod_{v\in c(r;r)} \underbrace{p(x_v|x_r) \prod_{u\in T_v\backslash \{v\}}p(x_u|x_{\pi(u;v)})}_{\text{Gaussian tree } T_v},
\end{eqnarray*}
so that if we observe a realization of the random variable $x_r$, we can compute its likelihood with $p(x_r)$, and we are left with a forest of Gaussian trees (note that $p(x_v|x_r)$ is a Gaussian distribution and we have $\pi(u;v)=\pi(u;r)$ for $v\in c(r;r)$ so that the (other) conditional probabilities are the same as in \eqref{eq:def_gt}).

In a Gaussian tree, it is possible to marginalize easily at any $v\in V$ thanks to the conjugacy properties of the Gaussian:
\begin{proposition}\label{prop:bp}
    Given a Gaussian tree $T$ with root $r$ and a neighbor of the root denoted $r'$, we have
    \begin{eqnarray*}
        p((x_v)_{v\in V}) = p(x_{r'}) \prod_{v\in V\backslash \{r'\}}p(x_v|x_{\pi(v;r')}),
    \end{eqnarray*}
    where $p(x_{r'}) =$\\
    $
         \mathcal{N}\left(x_{r'}| A_{(r'|r)}\mu_r+b_{(r'|r)},\Sigma_{(r'|r)}+A_{(r'|r)}\Sigma_r A_{(r'|r)}^T\right)
    $
    \\\\
    and for $v\neq r,r'$, we have $p(x_v|x_{\pi(v;r')})=p(x_v|x_{\pi(v;r)})$ and,
    $
        p(x_r|x_{r'}) = \mathcal{N}\left(x_r| A_{(r|r')}x_r+b_{(r|r')},\Sigma_{(r|r')}\right),
    $
    \\\\
    with $A_{(r|r')}$, $b_{(r|r')}$ and $\Sigma_{(r|r')}$ given by the standard conditional Gaussian distributions (see Appendix \ref{app:formulas}).
\end{proposition}
Our algorithm to marginalize for a rooted Gaussian tree at any vertex $v\in V$ is now straightforward: thanks to Proposition \ref{prop:bp} applied on the path from the root $r$ to $v$, compute the joint probability \eqref{eq:def_gt} with $v$ as the new root; after marginalization at the new root $v$, we obtain a forest of new rooted Gaussian trees so that we can iterate marginalization with the same procedure on each of them. 
\begin{remark}
    A more classic presentation of belief propagation consists in marginalizing all nodes thanks to a message-passing algorithm on the tree. Here, we compute the message passing only on the required path from the old root to the new one.
\end{remark}

\subsection{Rao-Blackwellized sequential Monte Carlo}\label{Sec:rao_smc}
We first show how our algorithm allows us to recover the Kalman filter. We consider the very simple Hidden Markov Model (HMM) given by: $x_0$ is a Gaussian r.v. and for $t\geq 0$,
\begin{eqnarray*}
    x_{t+1} = x_t + \epsilon^x_t,\text{ and }
    y_{t+1} = x_{t+1} +\epsilon^y_t,
\end{eqnarray*}
where $\epsilon^x_t$ and $\epsilon^y_t$ are independent Gaussian r.v.
Clearly for any $T>1$, the law of $(x_t,y_t)_{t\leq T}$ is given by a Gaussian tree where all the $x_t$'s are connected through a line when $t$ is increasing and the $y_t$ are leaves connected to the corresponding $x_t$. In the typical setting for the Kalman filter, we observe the $y_t$ and estimate the corresponding state $x_{t}$. Applying belief propagation on the Gaussian tree, when we observe $y_{t}$, we move the root to $y_t$, and marginalize it. Since the root where we marginalize is a leaf connected to $x_{t}$, we obtain a new Gaussian tree rooted at $x_{t}$ with an explicit Gaussian distribution for $x_{t}$. When we observe $y_{t+1}$, we can run the same algorithm. It is easy to check that we obtain the same analytic expression as the standard Kalman filter (and the extension to a linear model instead of the simple random walk model presented here is straightforward). 

Note that as $t$ increases, we have a larger and larger Gaussian tree (indeed a line in this case). Indeed, if we kept this growing tree, we could implement a smoothing algorithm to compute exactly the distribution $p(x_0,x_1,\dots, x_t|y_1,\dots, y_t)$ (but the memory requirement would grow with $t$). Here we are interested in filtering i.e., computing the distribution $p(x_t|y_1,\dots, y_t)$. In this case, since $(x_t)_t$ is a Markov chain, we can compute $p(x_{t+1}|y_1,\dots, y_{t+1})$ from $p(x_t|y_1,\dots, y_t)$ and ignore all the tree structure involving $x_0,x_1,\dots, x_{t-1}$. In our algorithm, once the tree is rooted at $x_t$, we never access the nodes corresponding to $x_0,\dots, x_{t-1}$ so that we can indeed remove them. In our implementation, this will be done automatically, thanks to the garbage collector of Julia, ensuring a bounded memory footprint.

We now consider the model presented in Section \ref{sec:ex_code}, where symbolic and numeric computations are done. The initial speed $s_0$ and position $x_0$ of the runner are Gaussian r.v. and the model is given by:
\begin{eqnarray}
\label{eq:model}
    s_{t+1} &=& s_t + \epsilon^s_t\\
\label{eq:model2}    s^{\text{obs}}_{t+1}&=& s_{t+1} +\epsilon^{\text{obs},s}_t\\
    x_{t+1} &=& x_t+s_t\\
    a_{t+1} &=& \text{alt}(x_{t+1})\\
    a^{\text{obs}}_{t+1} &=& a_{t+1} +\epsilon^{\text{obs},a}_t,
\end{eqnarray}
where $s_t$ is the speed of the runner, $x_t$ his position on the trail and $a_t$ his corresponding altitude. The variables with superscript obs are noisy measurements of the speed and altitude. We assume that the $\text{alt}$ function (giving the altitude) is known (but not linear).

This SSM is non-linear, and there is no tractable formula to compute estimates for the state (position, speed, and altitude). As a result, our algorithm will rely on sampling if this model is coded as presented in Section \ref{sec:ex_code}. But, we can use Remark \ref{re:obs} and "mix" the model and the inference by inserting observations inside the model: contrary to the code, we included an observation of the speed before computing the new position (which is possible within our framework).
As a result, we see that equations \eqref{eq:model} and \eqref{eq:model2} correspond exactly to the Kalman filter described previously. By ignoring the altitude observation, we have a Gaussian linear model for the speed for which computations can be made analytically. As a result, our algorithm will compute exactly $p(s_t|s^{\text{obs}}_t)$ and rely on sampling for the remaining $x_t$ and $a_t$. Our experiments below, show that this approach is much more efficient than a standard SMC.

\section{Experiments}\label{sec:expe} 
For our first experiment, we checked that our algorithm recovered the Kalman filter for linear SSM. To do so, we compared our numerical values with two Julia packages: \textit{Kalman.jl}\footnote{https://github.com/mschauer/Kalman.jl} (specifically built to run Kalman filter) and \textit{ReactiveMP.jl}\footnote{https://github.com/biaslab/ReactiveMP.jl which has now been subsumed into https://github.com/biaslab/RxInfer.jl} (an efficient reactive message passing based variational inference engine). We found that all three methods agree numerically. We then compared the execution times and found (see Figure \ref{fig:linearssm} in Appendix \ref{app:exp}) that the package built specifically for Kalman filter is the fastest, and that our algorithm is slower but can handle much more complex models.

\begin{figure}[ht]
\begin{center}
\centerline{\includegraphics[width=\columnwidth]{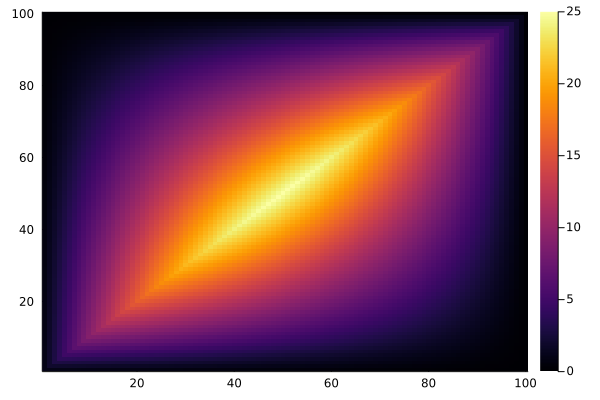}}
\caption{Covariance matrix computed by our Gaussian belief propagation for a (discretized) Brownian bridge.}
\label{fig:bridge}
\end{center}
\vskip -0.2in
\end{figure}

In our second experiment, we checked that our algorithm could mix analytical computation and observations (see Remark \ref{re:obs}): we modeled a simple random walk starting at zero and conditioned it to finish at time $T$ at zero by adding an observation for the last value of the random walk. We know that when the step size goes to 0, the limit is a Brownian bridge $B(t)$ with covariance structure given by $\text{Cov}(B(s),B(t)) = \min(s,t)-st/T$. We checked that our algorithm computed this covariance matrix (Figure \ref{fig:bridge}).

\begin{figure}[ht]
\vskip -0.2in
\begin{center}
\centerline{\includegraphics[width=\columnwidth]{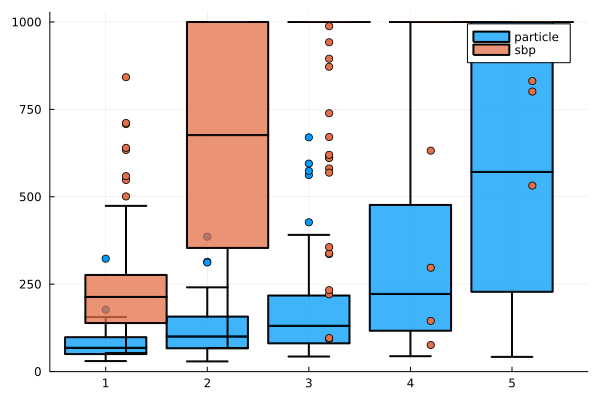}}
\caption{Diverging times for regular SMC (particle) and our algorithm (SBP) as a function of the number of particles with $2,5,10,20,40$ particles.}
\label{fig:divertimes}
\end{center}
\vskip -0.4in
\end{figure}

In our third experiment, we implemented the tracker model presented above. We simulate a runner to generate the observations and to make the problem more interesting, observations are only available every five timesteps. We generate runs of length 5000, and for each simulation, the measure of performance of the inference algorithm is the time at which the tracker diverges i.e., the runner is far away from the estimated position. Hence the longer this time is, the better the tracker. Figure \ref{fig:divertimes} shows the results comparing standard SMC (particle) and our algorithm sbp as explained in Section \ref{Sec:rao_smc}. Each column is a boxplot for the times of divergence of the algorithms over 100 simulations and for different numbers of particles used for sampling: column 1 with 2 particles, then 5, 10,20,40 particles. We see that even with 40 particles, the standard SMC diverges before the end of the run. However, already with 10 particles, our algorithm is tracking the runner until the end of the run for almost all simulations.

\section{Related work}\label{sec:related}

Our symbolic belief propagation method bears resemblance to delayed sampling (DS) \cite{delayed_sampling}. This algorithm also attempts to perform most computations symbolically. However, it maintains chains instead of trees and cannot invert parent-children relationships. As a consequence, it sometimes has to sample random variables that our algorithm would not and thus be less precise.
More recently, semi-symbolic inference~\cite{semi-symbolic} is a generalization of delayed sampling that is able to perform exact computations on closed families of distributions (e.g., linear Gaussian, or finite discrete models) at the cost of an increased overhead.
While less general, our solution is based on a well-known, efficient algorithm which can already compute the exact solution on a large class of models.

Our domain-specific language revolves around reactive probabilistic programming, which was introduced with ProbZelus \cite{probzelus}. A major difference is that our library integrates itself fully with the Julia ecosystem: models can both contain arbitrary Julia code and be called by any Julia code. To achieve this, we rely on and leverage the extensive metaprogramming abilities of Julia and in particular, its macro system. An interesting avenue for future work is the deeper integration of our framework with other metaprogramming libraries in Julia, such as automatic differentiation tools \citep{zygote}.

\bibliography{onlinesampling}
\bibliographystyle{icml2023}

\section{Appendix}
\subsection{Technical Formulas}\label{app:formulas}
\begin{proposition}
    Given a Gaussian tree $T$ with root $r$ and a neighbor of the root denoted $r'$, we have
    \begin{eqnarray*}
        p((x_v)_{v\in V}) = p(x_{r'}) \prod_{v\in V\backslash \{r'\}}p(x_v|x_{\pi(v;r')}),
    \end{eqnarray*}
    where $p(x_{r'}) =
         \mathcal{N}\left(x_{r'}| A_{(r'|r)}\mu_r+b_{(r'|r)},\Sigma_{(r'|r)}+A_{(r'|r)}\Sigma_r A_{(r'|r)}^T\right)
    $
    \\and for $v\neq r,r'$, we have $p(x_v|x_{\pi(v;r')})=p(x_v|x_{\pi(v;r)})$ and,
        $p(x_r|x_{r'}) = \mathcal{N}\left(x_r| A_{(r|r')}x_r+b_{(r|r')},\Sigma_{(r|r')}\right)$,
    with $A_{(r|r')}$, $b_{(r|r')}$ and $\Sigma_{(r|r')}$ given by the standard conditional Gaussian distributions:
    \begin{eqnarray*}
        A_{(r|r')} &=& \Sigma_{(r|r')}\left(A_{(r'|r)}^T\Sigma_{(r'|r)}^{-1} \right),\\
        b_{(r|r')} &=& \Sigma_{(r|r')}\left(\Sigma_r^{-1}\mu_r-A_{(r'|r)}^T\Sigma_{(r'|r)}^{-1}b_{(r'|r)} \right),\\
        \Sigma_{(r|r')} &=& \left( \Sigma_r^{-1} + A_{(r'|r)}^T\Sigma_{(r'|r)}^{-1}A_{(r'|r)}\right)^{-1}.
    \end{eqnarray*}
\end{proposition}

\subsection{More experiments}\label{app:exp}
\begin{figure}[ht]
\vskip 0.2in
\begin{center}
\centerline{\includegraphics[width=\columnwidth]{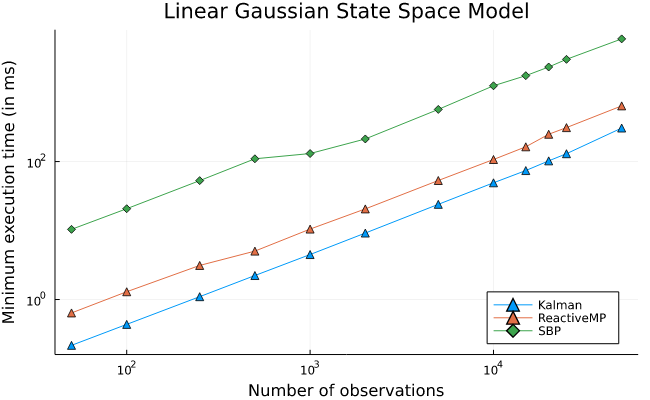}}
\caption{Execution times as a function of the number of observations. Comparison of our algorithm SBP (Stream belief propagation) and \textit{Kalman.jl} and \textit{ReactiveMP.jl}}
\label{fig:linearssm}
\end{center}
\vskip -0.2in
\end{figure}
\end{document}